\documentclass[runningheads]{llncs}
\usepackage{graphicx}
\usepackage{amsmath,amssymb} 
\usepackage{color}

\usepackage{times}
\usepackage{epsfig}
\usepackage{epstopdf}
\usepackage{multirow}
\usepackage{setspace}
\usepackage{enumitem}

\usepackage[ruled,linesnumbered]{algorithm2e}
\usepackage{array}

\let\oldnl\nl
\newcommand{\nonl}{\renewcommand{\nl}{\let\nl\oldnl}}

\newlength\savewidth

\newcolumntype{C}{>{\centering\arraybackslash} m{3.9cm} }

\usepackage{CJK}

\usepackage[pagebackref=true,breaklinks=true,letterpaper=true,colorlinks,bookmarks=false]{hyperref}

\begin{document}
\begin{CJK*}{UTF8}{song}
\pagestyle{headings}
\mainmatter


\title{Better Image Segmentation by Exploiting Dense Semantic Predictions} 
\titlerunning{Better Image Segmentation}
\authorrunning{Qiyang Zhao, Lewis D Griffin}

\author{Qiyang Zhao, Lewis D Griffin}
\institute{Beihang University \& UCL}

\maketitle

\begin{abstract}

It is well accepted that image segmentation can benefit from utilizing multilevel cues. The paper focuses on utilizing the FCNN-based dense semantic predictions in the bottom-up image segmentation, arguing to take semantic cues into account from the very beginning. By this we can avoid merging regions of similar appearance but distinct semantic categories as possible. The semantic inefficiency problem is handled. We also propose a straightforward way to use the contour cues to suppress the noise in multilevel cues, thus to improve the segmentation robustness. The evaluation on the BSDS500 shows that we obtain the competitive region and boundary performance. Furthermore, since all individual regions can be assigned with appropriate semantic labels during the computation, we are capable of extracting the adjusted semantic segmentations. The experiment on Pascal VOC 2012 shows our improvement to the original semantic segmentations which derives directly from the dense predictions.

\end{abstract}

\section{Introduction}

People have realized the importance of utilizing multilevel cues in image segmentation for quite long a time \cite{Martin2001}\cite{Cremers}. The usual cues we can exploit are appearances in the low level, contours in the mid-level and semantic cues in the high level. In the widely-adopted image segmentation methods, the low and mid- level cues are exploited more frequently. It is sensible as the segmentation task is to partition images into regions of uniform perceptual features \cite{Bagon}. Segmentation methods developed in light of graph theory \cite{Felzenszwalb}, nonparametric clustering \cite{Comaniciu}, information theory \cite{Ma}, correlation clustering \cite{Yarkony} and manifold embedding \cite{Yu}, exploit the appearance cues in diverse ways. Recent researches attempt to learn mid-level cues such as contours from human segmentations \cite{Dollar1}\cite{Donoser}\cite{Martin}\cite{RenX3}\cite{Xie}\cite{Kokkinos}, then transform contour maps into segmentation hierarchies using the Ultrametric Contour Map (UCM) method \cite{Arbelaez2}\cite{Arbelaez1}\cite{Arbelaez3}\cite{Ren}. Rather than learning from only the final human segmentations, it is suggested in \cite{Zhao} to learn from human behaviors conducted in the annotation processes.

However, some mistakes might be made if only appearance and contour cues are exploited, because it is hard to discriminate the regions of similar appearances but distinct semantic categories. A good example is the animals of protective colorations in the wild fields. Therefore there are a number of literatures arguing that the semantic cues should be exploited \cite{Cremers}\cite{Maire1}\cite{Martin}, optionally as the high-level features \cite{Russell}\cite{Tu}\cite{Wu}. It should be noticed that although multilevel cues are involved in these approaches, the cues are utilized asynchronously rather than simultaneously. To be detailed, image pixels are grouped into superpixels according to their appearances, then the specific models are established to exploit semantic cues \cite{Borenstein}\cite{Wu}.The advantage is, people can adopt different well-developed tools such as conditional random fields (CRF) in individual stages to develop segmentation algorithms easily. However as discussed in Sec. \ref{sec:integratingSemanticCues}, postponing to exploit semantic cues can hardly avoid the similar-apperance-distinct-category mistakes though. An exception is \cite{Maire1} which uses a joint embedding of semantic cues and appearance cues, but there we have to solve the Normalized Cuts \cite{Shi} with the huge computational load, and it is not flexible enough to accommodate the emerging dense semantic cues.

\begin{figure*}[htbp]
	\centering
	\includegraphics[width=1.0\linewidth]{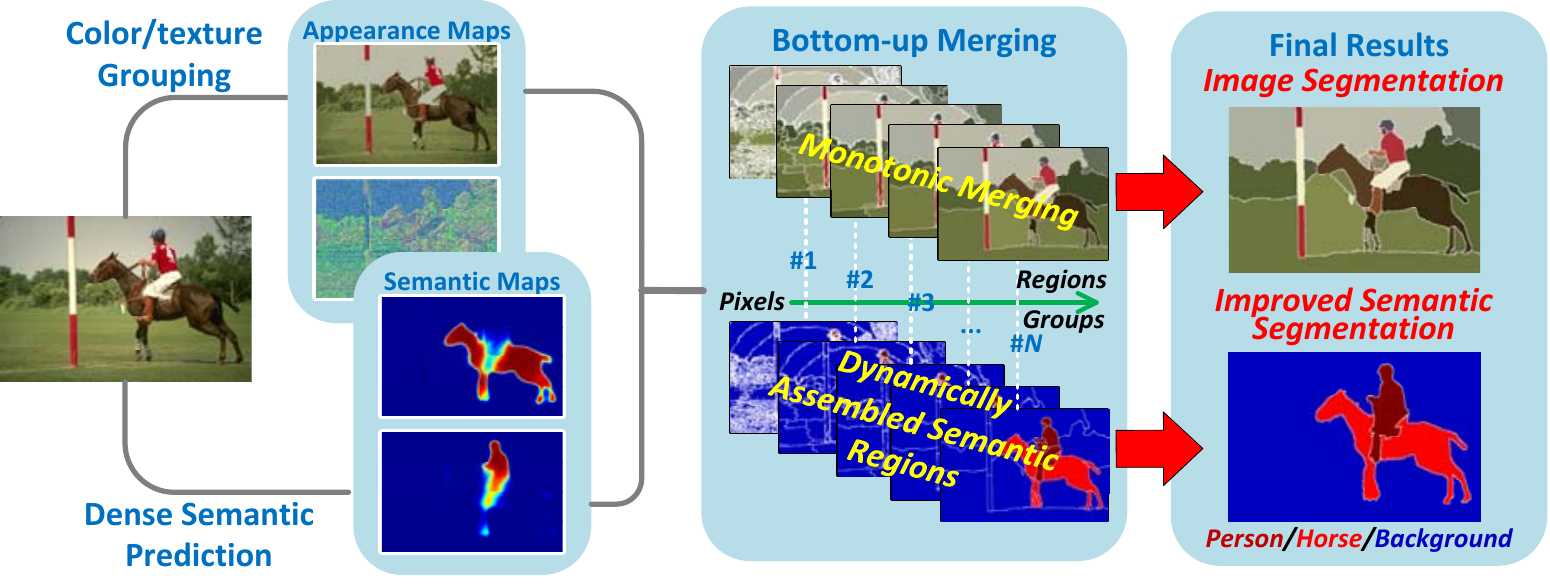}
	\caption{The pipeline. Given an image, appearance maps and semantic maps are produced to calculate region description complexities. The bottom-up segmentation process generates a segmentation hierarchy together with a sequence of varying category-wise groups. Finally by thresholding we obtain the image segmentation, optionally with the adjusted semantic segmentation.}
	\label{fig:flowchart}
\end{figure*}

The semantic segmentation methods can produce semantic cues of fine granularities for arbitrary images. The precursor approaches obtain category-wise groups by refining object detection boxes \cite{Brox}\cite{Parkhi}\cite{Yang}, or classifying superxiels \cite{Arbelaez1} and object proposals \cite{Carreira}\cite{Uijlings} into different semantic categories \cite{Arandjelovic}\cite{Ion}. They deal with a severely limited number of semantic categories, and the segmentation precisions are usually uncompetitive. Fortunately, people have made significant progresses based on Fully-Convolutional Neural Network (FCNN) \cite{Long} in recent years. The FCNN-based semantic segmentations can be improved further \cite{Chen}\cite{Dai}\cite{Lin}\cite{Zheng} using existing tools such as fully connected conditional random fields \cite{Krahenbul}. The joint training of FCNN together with CRF will bring more benefit to the precisions \cite{Liu}. The FCNN-based methods are of the state-of-the-art performance, thus more qualified as a reliable source of semantic cues. However so far as we know, few image segmentation approaches are developed to exploit the emerging FCNN-based dense semantic predictions. 

In Sec. \ref{sec:integratingSemanticCues}, we propose to utilize the multilevel cues including the dense FCNN-based semantic cues, in the framework of \cite{Zhao}. Semantic cues are taken into account from the very beginning of the region merging processes. Since there are usually unexplored categories beyond the capacity of even FCNN-based semantic models, we propose an approximation method to handle the semantic inefficiency problem in Sec. \ref{sec:insufficiency}. Besides this, the robustness of multilevel cues is suggested to play a key role in the quality of image segmentations \cite{Yu}\cite{Fu}, thus we propose a straightforward way in Sec. \ref{sec:noise} to use contour cues to suppress the noise in multilevel cues. Finally we perform experiments to test our new methods on BSDS500 \cite{Martin2001} and Pascal VOC12 \cite{Everingham} in Sec. \ref{sec:exp}.

\section{Related Work}
\label{sec:relatedwork}

\begin{figure}[htbp]
	\centering
	\begin{tabular}{cccc}
		{\includegraphics[width=2.8cm]{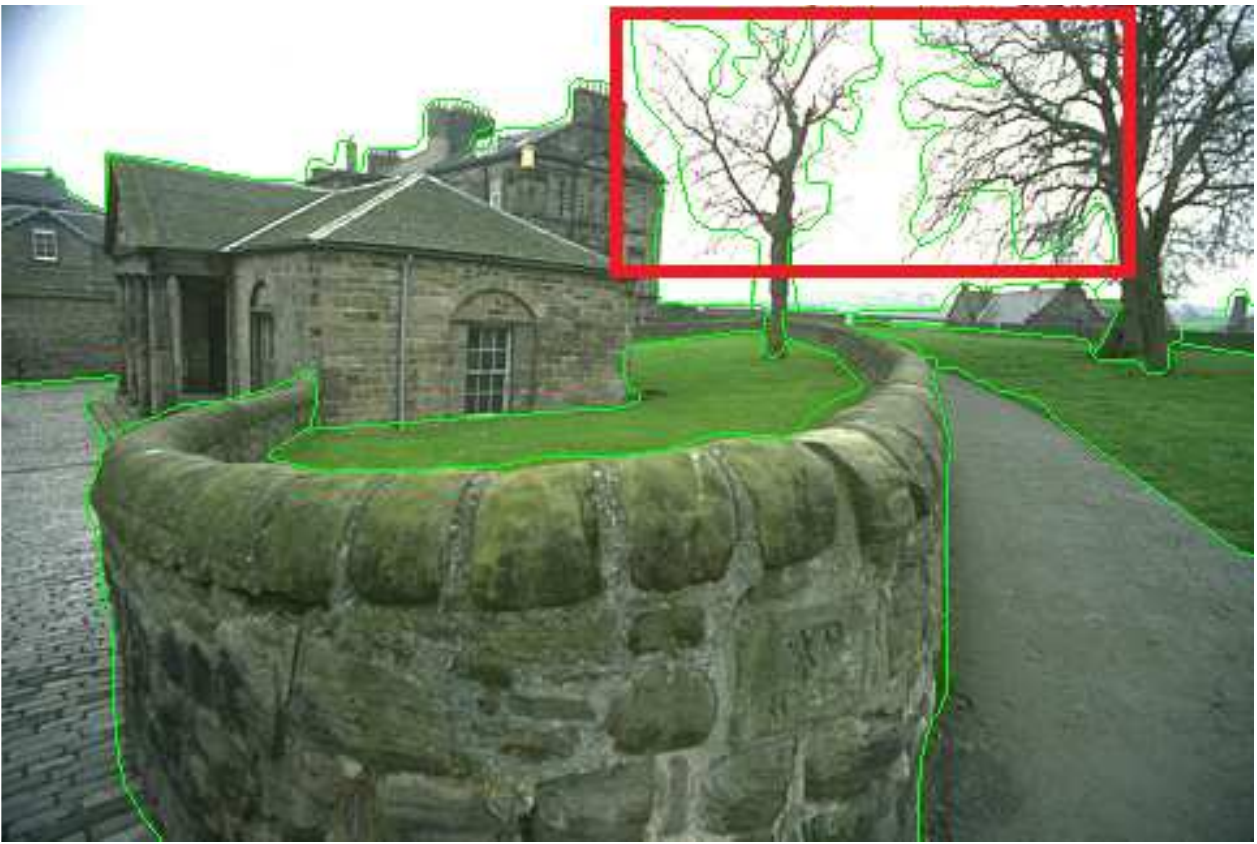}}&
		{\includegraphics[width=2.8cm]{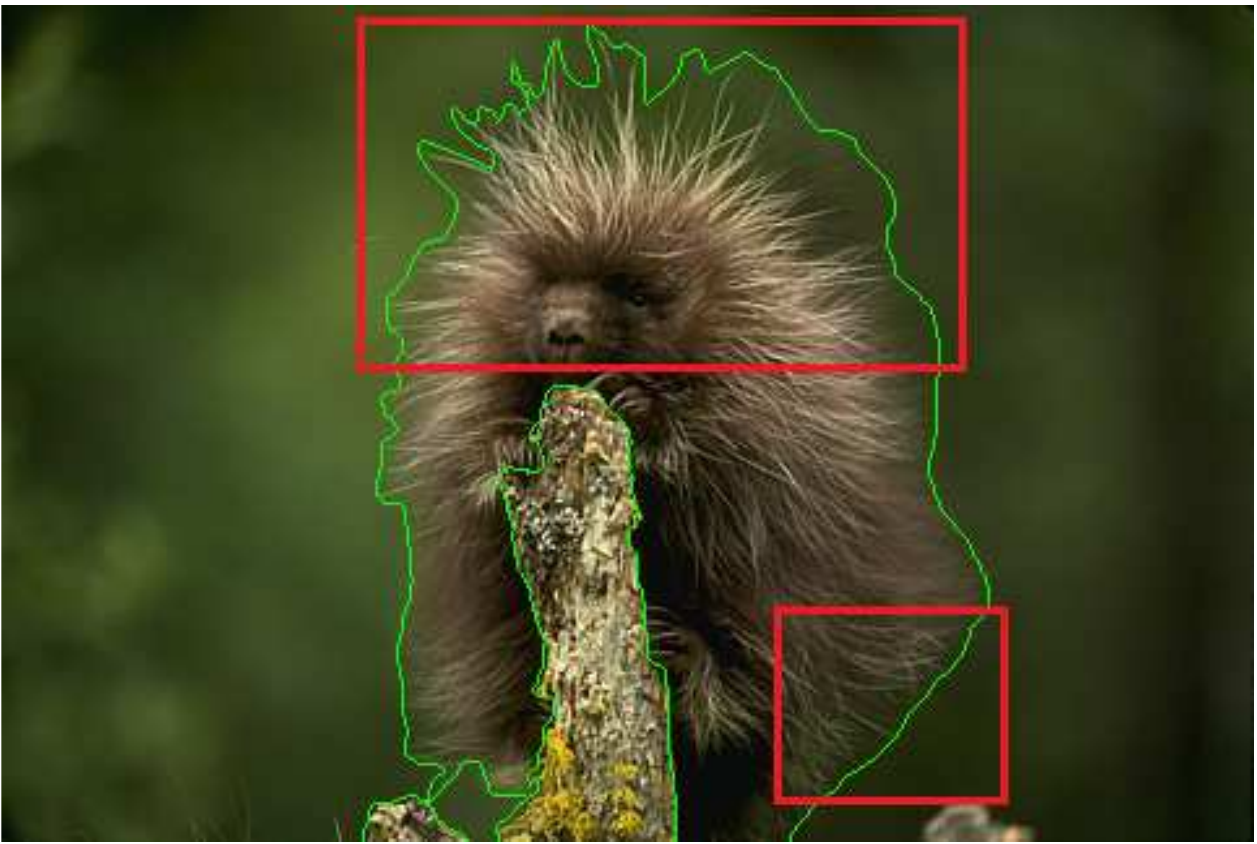}}&
		{\includegraphics[width=2.8cm]{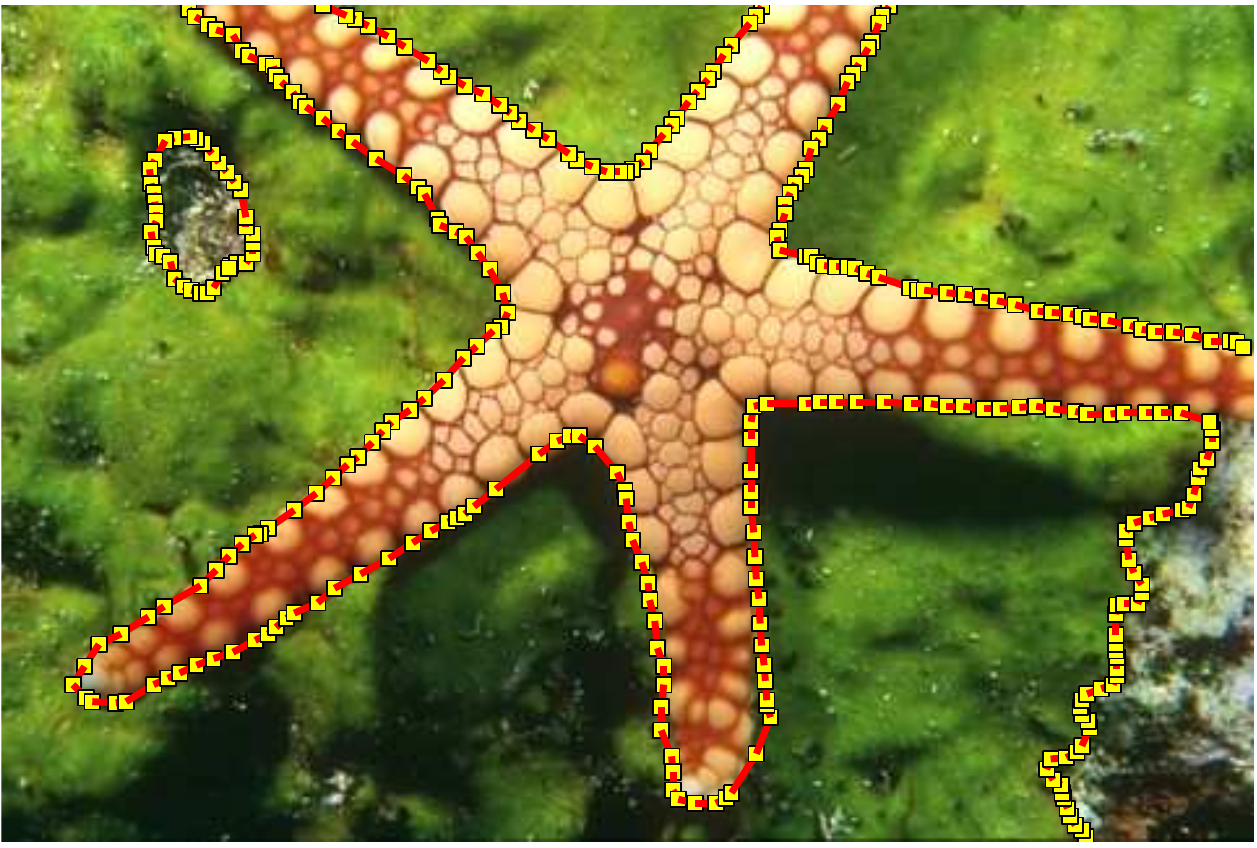}}&
		{\includegraphics[width=2.8cm]{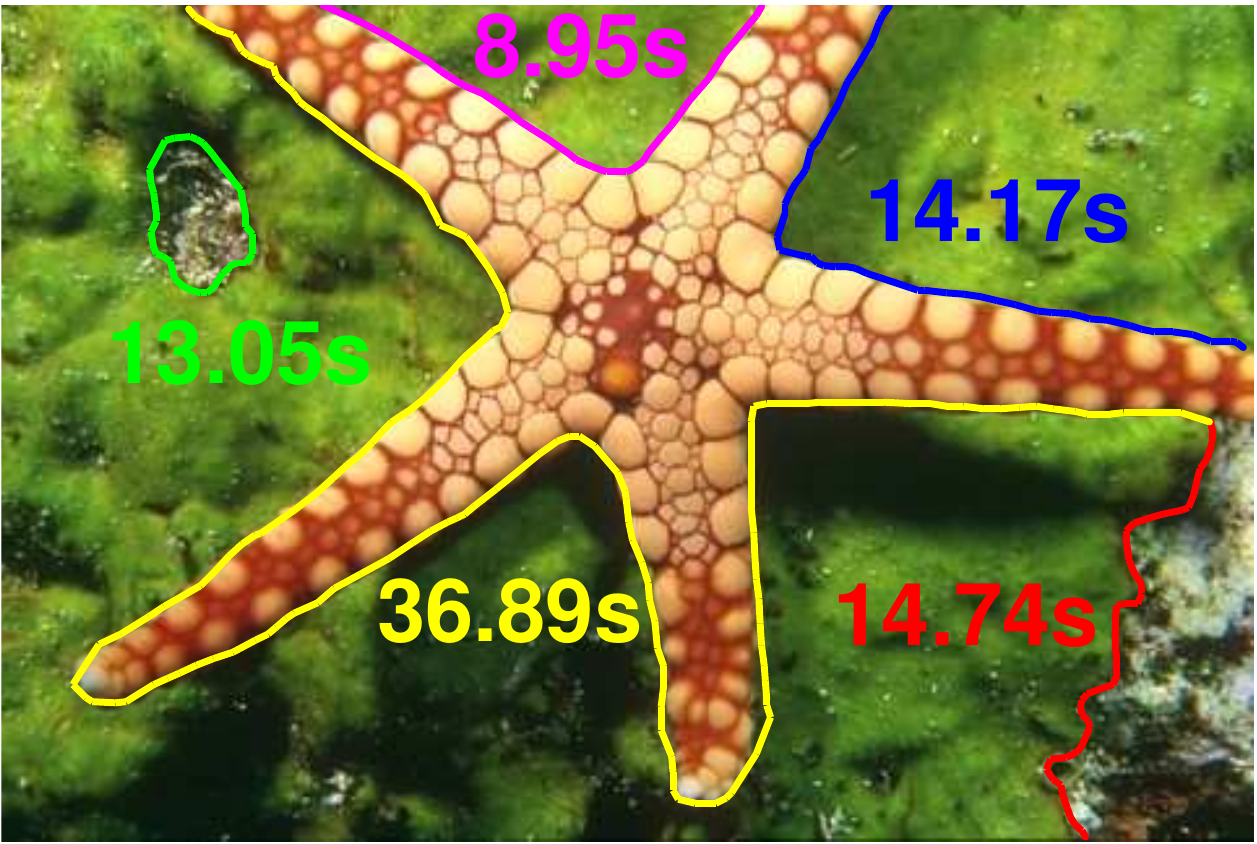}}\\
		(a)&(b)&(c)&(d)
	\end{tabular}
	\caption{Boundary mistakes in human segmentations and the tracing experiments. In (a)(b), boundaries are shown in green and the inaccurate parts are bounded by red boxes; (c) tracing paths marked by mouse clicks of small yellow squares; (d) boundaries traced in (c) and time consumptions in seconds (in the same colors). The images and data are from \cite{Zhao}.}
	\label{fig:boundarytracing}
\end{figure}

In the bottom-up merging segmentation approaches \cite{Arbelaez1}\cite{Arbelaez3}, it is beneficial to use contour cues learned from human annotations. Then rather than treating human segmentations as only static samples, what other stuff can we learn? In \cite{Zhao}, it is noticed that although human segmentations are of the high quality, there are surprisingly many boundary mistakes, even in the renown BSDS500 dataset, see Fig. \ref{fig:boundarytracing}.a-b. Easy to see, these mistakes are made intentionally instead of by casual operations. 

It reminds us there are some subtle facts which are not discovered in the human annotations. Here it should be clarified that we do not aim to teach machines reproduce boundary mistakes, but are interested in the underlying principles which bring us high quality segmentations. The above mistakes are just the by-products. It is argued in \cite{Zhao} that, the human annotation processes are subject to \emph{the least effort principle} (LEP) which governs a wide range of human behaviors: a human \emph{will strive to solve his problem in such a way as to minimize the total work that he must expend} \cite{Zipf}. When annotating segmentations, the effort includes understanding images in brains and figuring out the boundaries by input devices. The former is hard to model, thus only the later one is investigated in \cite{Zhao}. When tracing the boundaries, human subjects are inclined to choose simpler paths to save their effort, such as in Fig. \ref{fig:boundarytracing}.a-b. The tracing effort $T$ is considered to be the product of (subjective) \emph{unit tracing cost} $\lambda$ and the tracing load $L$ which is measured by time consumptions. Based on the observations in boundary tracing experiments supported by the annotation tool \cite{BSDStool} (see Fig. \ref{fig:boundarytracing}.c-d), the tracing load are estimated by the mixture of experts in \cite{Zhao}.

Here the problem human subjects need solve is to partition images into regions of single things \cite{Martin2001}. This target can be formulated in terms of region description complexities: the lower the complexity is, the more likely only one single thing is in a region \cite{Bagon}\cite{Rao}. However it is ambiguous though for the case of moderate complexities. Generally, there is a decrease of the total description complexity if a region is partitioned into sub-regions. It is argued in \cite{Zhao} we should partition a region if and only if the extra effort we need make is less than the complexity decrease. A \emph{singleness} predicate $sngl$ is proposed to indicate whether a region $R$ should be retained rather than partitioned
\begin{equation}
	\label{eq:retainable}
	sngl(R)=\left[\forall \{R_i\}, D(R)-\sum{D(R_i)}\le T(\{R_i\})\right]
\end{equation}
where $\{R_i\}$ is a partition configuration of $R$, $D(\cdot)$ is the region description complexity, and $T$ is the effort to trace boundaries between the regions in $\{R_i\}$. According to LEP, a human subject will finish his segmentation task by solving
\begin{equation}
	\label{eq:minL}
	\min_{\mathcal{S}}\ \lambda L\left(\mathcal{S}\right),\text{s.t.}\ \forall\ R \in \mathcal{S}, sngl(R)=\emph{\textbf{TRUE}}.
\end{equation}
where $\mathcal{S}$ is the segmentation configuration in terms of a set of regions, and $L\left(\mathcal{S}\right)$ is the total tracing load of all boundaries induced by $\mathcal{S}$. The paper uses a bottom-up merging process to inversely simulate the top-down human segmentation process \cite{Zhao}, and the corresponding merging criterion is called the \emph{unit merging cost}\footnote{The original term is \emph{merging unit cost} in \cite{Zhao}. Here we change the order of the terms slightly for easier understanding.}
\begin{equation}
	\label{eq:umc}
	umc(\{R_i\})=\frac{D(\cup_i R_i)-\sum_i{D(R_i)}}{L(\{R_i\})}.
\end{equation}

The paper solves a restricted version of eq. \ref{eq:minL} under the \emph{hierarchy} and \emph{monotonicity} constraints, by repeatedly merging a pair of regions of the currently minimum \emph{umc}. The corresponding segmentation algorithm exhibits the good boundary and region performance, together with the high efficiency.

However, it is not well-investigated in \cite{Zhao} how to accommodate the multilevel cues in the merging criteria, not to mention the emerging dense semantic predictions. The Fully Convolutional Neural Networks (FCNN) are proposed in \cite{Long}, then widely used in semantic segmentation tasks. The FCNNs take the pixels as the input, and follow the basic structures of the standard convolutional neural networks which succeed in image classification challenges. Furthermore, the FCNNs can deal with images of arbitrary sizes, and gain flexible granularities using multiple processing streams \cite{Long}. The corresponding semantic segmentation methods can produce pixel-wise semantic labels, to say, maps of the probabilities of belonging to specific categories, see Fig. \ref{fig:flowchart}. They deal with the images without any preprocessing stages, and the results can be improved with the existing tools such as conditional random fields \cite{Chen}\cite{Lin}\cite{Zheng}. The FCNN-based methods obtain the state-of-the-art on several popular challenges such as the Pascal 2012 \cite{Everingham} and the MS COCO \cite{Lin-MS}.

Compared with the poor capacity of their precursors \cite{Maire1}\cite{Borenstein}, the FCNN-based approaches can deal with tens or even hundreds of categories. This meets the situation of image segmentation well, providing us the better support to handle the images in the wild. Nevertheless, it should be mentioned that the boundary accuracy is not an adherent objective for most semantic segmentation approaches, although the manually-annotated groundtruths are indeed composed of well-shaped regions. On the contrary, the region boundaries are emphasized in image segmentation tasks, thus it is interesting to observe whether we can benefit semantic segmentations on region boundaries in return.

\section{Utilizing Semantic Cues in Image Segmentation}
\label{sec:integratingSemanticCues}

\begin{figure}[htbp]
	\centering
	\includegraphics[width=9.5cm]{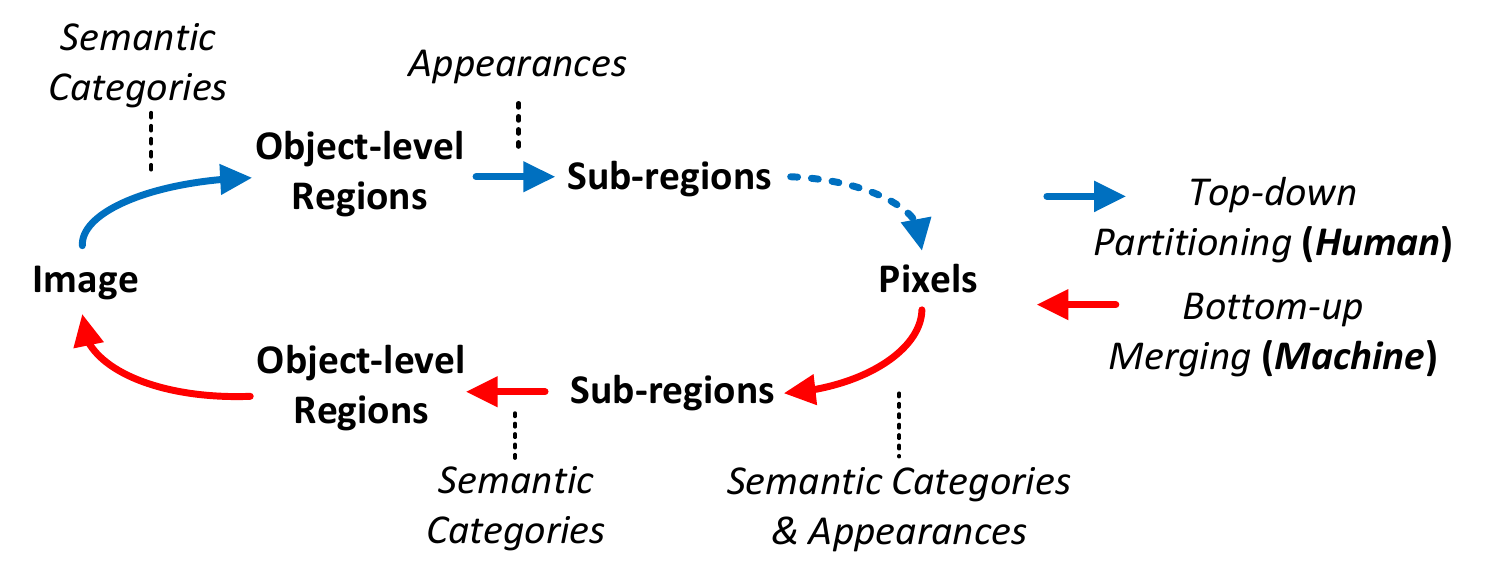}
	\caption{The major supportive cues in different stages. Semantic cues are not essential for humans to partition category-level regions into sub-regions of different appearances, while they are essential for machines to discriminate pixels of different categories when merging them into sub-regions. The stage of the dotted arrow in blue seldom happens, but we present it here for the completeness considerations.}
	\label{fig:2pipelines}
\end{figure}

In order to find out the role of semantic cues in image segmentation, we turn to observing the human annotation processes. A good example is the widely-adopted image segmentation dataset, BSDS500. It is observed that human subjects are used to partition images hierarchically in the top-down style \cite{Martin2001}\cite{BSDStool}: people partition images into large regions of distinct semantic categories, then divide them sequentially into sub-regions according to appearances, as shown in Fig. \ref{fig:2pipelines}. 

We use the bottom-up merging approach to simulate the human annotation processes inversely. In the final human segmentations, the pixels in each sub-region have two labels indeed: semantic category and appearance type. Consequently, it is only with both two labels the right way to discriminate pixels of different sub-regions. Therefore in the bottom-up machine processes, we need to use semantic cues at the very beginning, together with appearance cues to group pixels into appropriate sub-regions.

\begin{figure}[htbp]
	\centering
	\includegraphics[width=12.0cm]{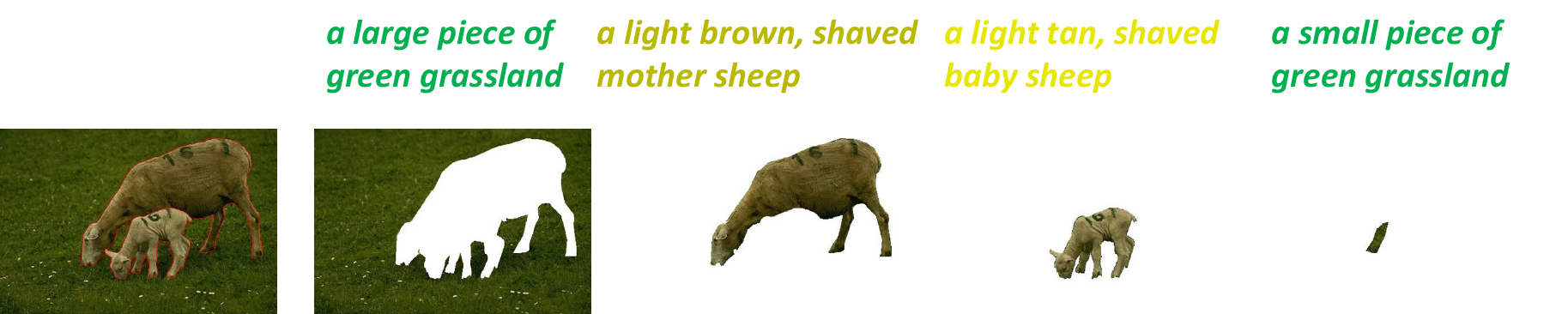}
	\caption{Image and pieces of single things. Left: the image and groundtruth boundaries. From middle left to the right: four pieces. The involved single things are described on the top of the pieces. Note: the image and groundtruth are from the BSDS500 train set.}
	\label{fig:singlethings}
\end{figure}

We notice that people are used to describe regions with the combinations of both appearances and semantic categories, such as \emph{a dark dog} and \emph{a curtain of black-white squares}, as shown in Fig. \ref{fig:singlethings}. It means the region description complexity is the sum of both appearance and semantic terms. We group colors and textures into clusters, then calculate discrete entropies as the color or texture description complexities for individual regions. A regularity term is designed for regions of slowly-varying appearances. It is the weighted sum of contour cues and regularized color distances between neighbor pixels. The regularized color distances is inspired by the probabilistic density of gradients \cite{Jia}. Then, we transform semantic cues into complexities and add them to the region description complexity $D(\cdot)$ in (\ref{eq:retainable})
\begin{equation}
	\label{eq:D}
	D(R)=w_cD_{c}(R)+w_tD_{t}(R)+w_sD_{s}(R)+w_rD_{r}(R)
\end{equation}
where $D_c,D_t,D_s,D_r$ are color, texture, semantic and regularity description complexities in turn, and $w_c,w_t,w_s,w_r$ are weights which need to be trained.

If semantic cues are accurate enough, we can directly pick up pixels according to their semantic labels to assemble category-wise groups. Then it seems there is not space to gain improvements with image segmentations in hand. However in practice, due to the huge difficulty of semantic prediction tasks, there is much noise in cues from even FCNN-based approaches, especially those nearby the region boundaries, such as the feet and tail in the horse map in Fig. \ref{fig:flowchart}. Low level cues such as colors and textures are more effective in locating the boundaries \cite{Zitnick}. Thus we hypothesize image segmentations are useful to correct mistakes in semantic cues to get accurate boundaries and consequently sensible regions. As a comparison, small superpixels can not provide equal support as they have lots of fake boundaries. 

\subsection{The Framework}

Here's the framework utilizing dense semantic cues in Alg. \ref{alg:unified framework}:

\begin{algorithm}\small
	\caption{Segmentation using Dense Semantic Cues}
	\label{alg:unified framework}
	\KwIn{Image $I$, semantic map $\mathcal{M}$ for labels in $\mathcal{L}$, threshold $\lambda_{\tau}$}
	\KwOut{Segmentation $\mathcal{S}$, category-wise groups $\{G_l\}$.}
	$S\leftarrow\{\langle p_i,l_i\rangle\}, G_l\leftarrow \phi$;  \textcolor[rgb]{0,0.5,0}{\text{//} each pixel $p_i$ as a region} \\
	\ForEach{$p_i\in I$}
	{
		$l_i\leftarrow\arg\min_{l} D_s(p_i|l,\mathcal{M})$; 		\textcolor[rgb]{0,0.5,0}{\text{//} initial label of pixel $p_i$}\\
		$G_{l_i}\leftarrow G_{l_i}\cup\{p_i\}$ \textcolor[rgb]{0,0.5,0}{\text{//} put pixels in initial groups for individual categories}\\
	}
	
	$\langle R_a,R_b\rangle\leftarrow\arg\min{umc(p_i,p_j)}, \lambda\leftarrow umc(R_a,R_b)$;\\
	\nonl \textcolor[rgb]{0,0.5,0}{\text{//} the minimum \emph{umc} on all  pairs of adjacent pixels}\\
	\While{$\lambda<\lambda_{\tau}$}
	{
		$R_{ab}\leftarrow R_a\cup R_b,l_{ab}\leftarrow \arg\min_{l} D_s(R_{ab}|l,\mathcal{M});$ \\
		$\mathcal{S}\leftarrow \mathcal{S}\cup\{ R_{ab}\}-\{ R_a,R_b\},G_{l_a}\leftarrow G_{l_a}-R_a,G_{l_b}\leftarrow G_{l_b}-R_b,$ \\
		\nonl $G_{l_{ab}}\leftarrow G_{l_{ab}}\cup R_{ab};$ \textcolor[rgb]{0,0.5,0}{\text{//} updating segmentation and category-wise groups} \\
		collect new corners and update related tracing loads;\\	
		update $umc(R,R_{ab})$ for each $R$ adjacent to $R_{ab}$ by (\ref{eq:handlinginsufficiency});\\
		$\langle R_a,R_b\rangle\leftarrow\arg\min{umc(R_i,R_j)},\lambda\leftarrow umc(R_a,R_b)$;\\
		\nonl \textcolor[rgb]{0,0.5,0}{\text{//} the minimum \emph{umc} on all  pairs of adjacent regions in $\mathcal{S}$}\\
	}
\end{algorithm}

Initially, each pixel is regarded as a region and put into a group according to the semantic map. Then we keep merging the region pairs of minimum $umc$'s in each loop until the unit merging cost reaches the threshold $\lambda_\tau$. Each time two region merge, the new merging corners are estimated and the corresponding tracing load is updated. We also use the postponed updating strategy in \cite{Zhao} to reduce wasted calculations in step 11 when updating $umc$'s. All $umc$'s and region pairs are stored in a direct access table \cite{Cormen}, so we can search, insert or remove a record in the constant time. Along with the region merging, all interested category-wise groups are dynamically assembled. Finally we obtain the image segmentation $\mathcal{S}$ together with groups $\{G_l\}$ of specific categories.

For each pixel $p$, the semantic map $\mathcal{M}$ stores a vector $\langle v_{p,1},v_{p,2},...,v_{p,N}\rangle$ measuring its memberships of semantic categories in $\mathcal{L}=\{l_1,l_2,\cdots,l_N\}$. We define the semantic description complexity of a region $R$ on category $l$ as 
\begin{equation}
	\label{eq:d_s_l}
	D_s(R|l,\mathcal{M})=-\sum_{p\in R}{\log_2{v_{p,l}}}.
\end{equation}
Each region has its most sensible semantic category. We let it be the category of the minimum description complexity
\begin{equation}
	\label{eq:bestsemlabel}
	l_R=\arg\min_{l} D_s(R|l,\mathcal{M}),
\end{equation}
and the semantic description complexity in (\ref{eq:D}) be
\begin{equation}
	\label{eq:d_s}
	D_s(R|\mathcal{M})=\min_{l} D_s(R|l,\mathcal{M}).
\end{equation}
We omit $\mathcal{M}$ in $D_s(\cdot)$ for simplicity in the below.

\section{Handling the Semantic Insufficiency}
\label{sec:insufficiency}

Most FCNN-based semantic models focus on hundreds of object categories at most, and all objects of other categories are treated as the \emph{background}. According to eq. \ref{eq:bestsemlabel} and \ref{eq:d_s}, $d_s$ defined on two \emph{background} regions is always zero, thus incapable of discriminating regions of unknown categories. We call it the semantic insufficiency problem.

\begin{figure}
	\centering
	\begin{tabular}{cccc}
		\includegraphics[width=2.7cm]{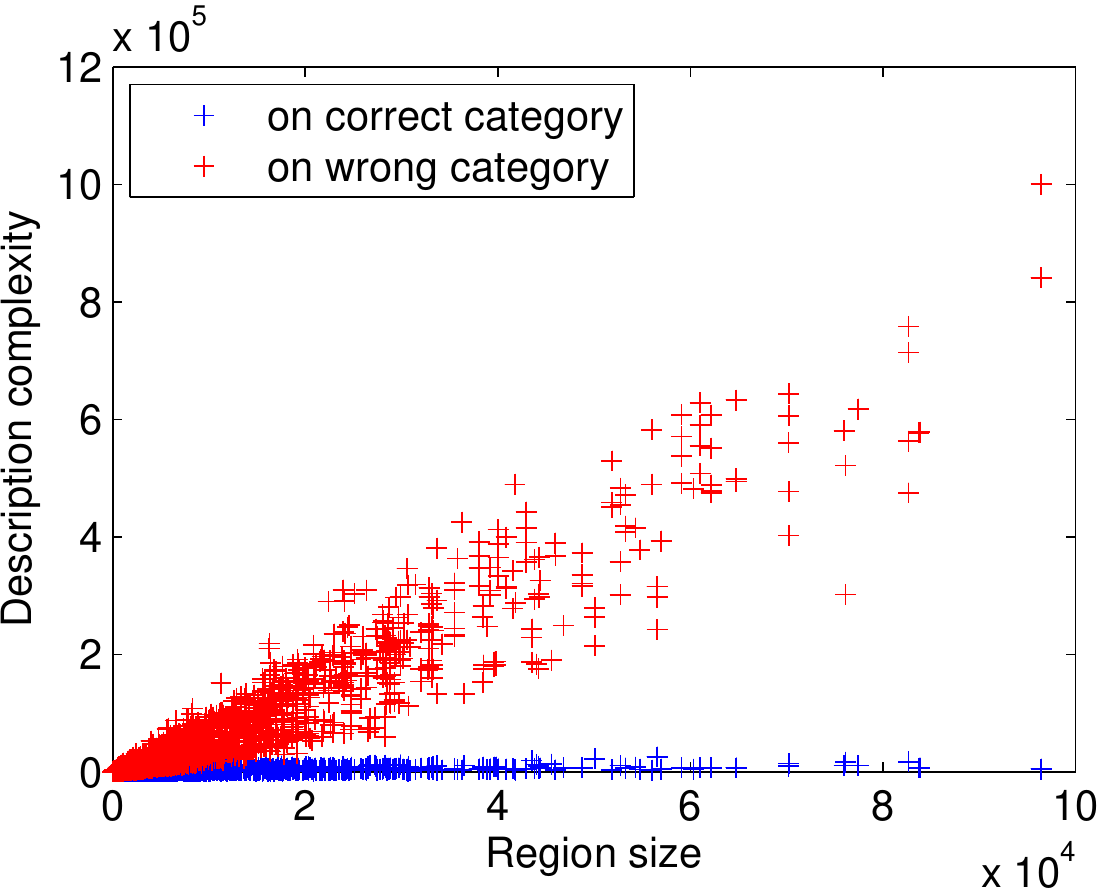}&
		\includegraphics[width=2.7cm]{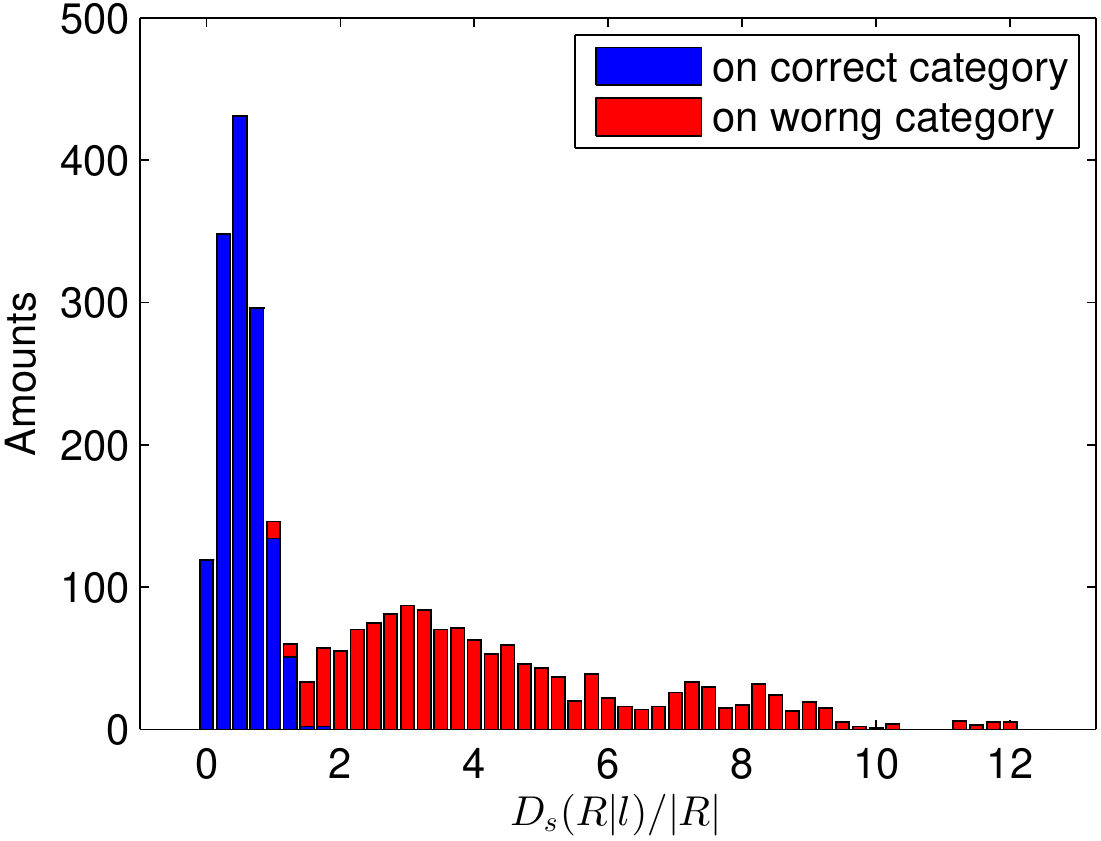}&
		\includegraphics[width=3.0cm]{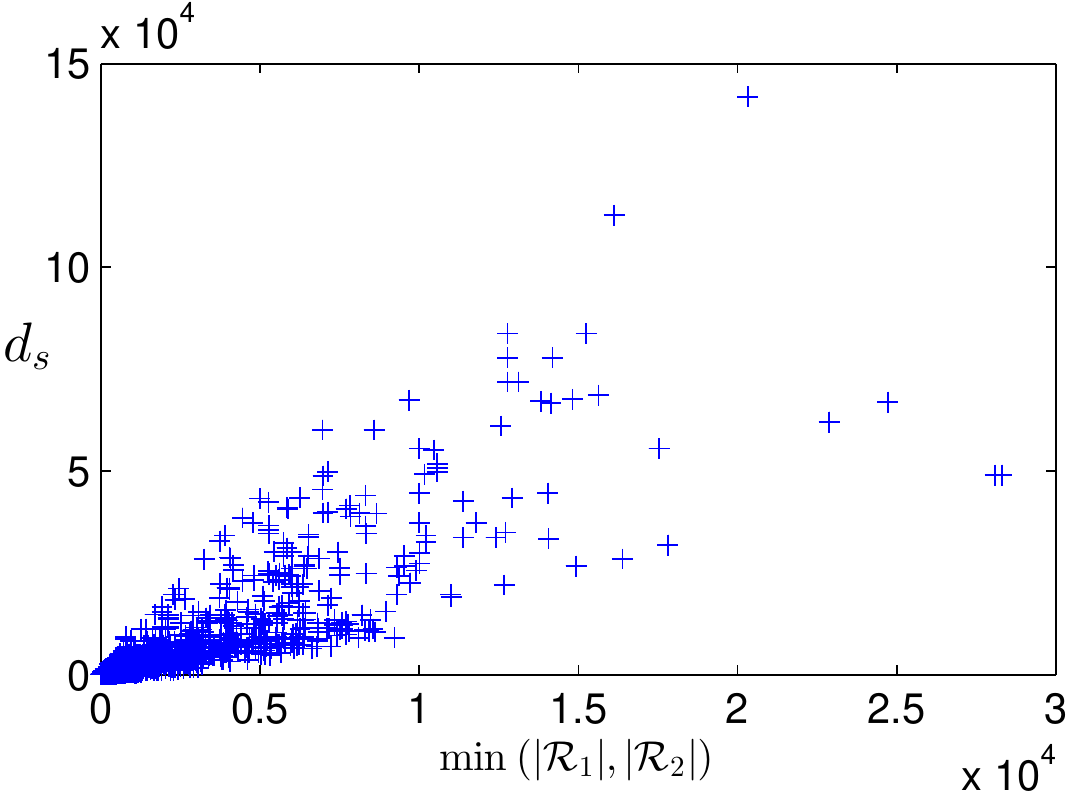}&
		\includegraphics[width=3.1cm]{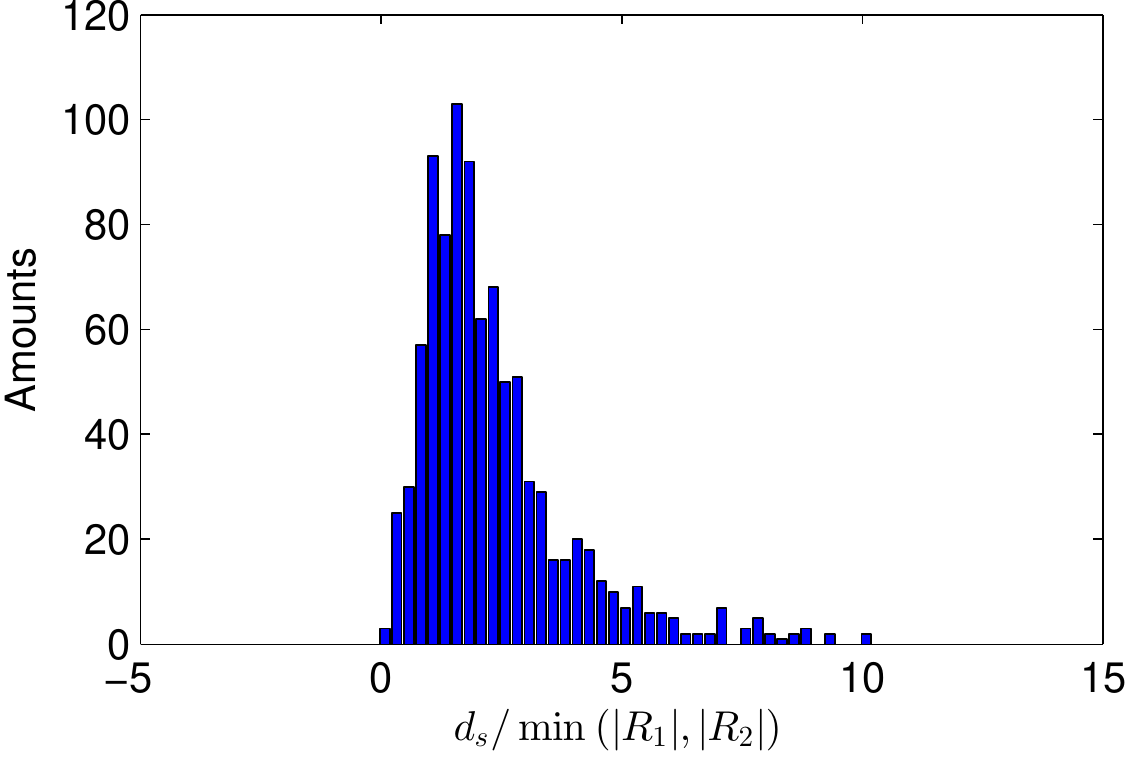}\\
		\ \ \ \ (a) & \ \ \ \ \ \ (b) & \ \ \ \ (c) & \ \ \ \ \ \ (d)\\
	\end{tabular}
	\caption{Semantic description complexities and region sizes. (a) $D_s(R|l)$ and region sizes. For each region $R$, we collect two samples by letting $l$ be correct and wrong categories respectively. (b) Histogram of the ratio of $D_s(R|l)$ to region sizes. (c) $d_s$ and the minimum region sizes. (d) Histogram of the ratio of $d_s$ to the minimum region sizes. Most ratios are close to 2.2. Note: All samples are collected from the BSDS500 train set. Zoom in for better views.}
	\label{fig:d_s_approx}
\end{figure}

For all pixels, their semantic description complexities should be much small on \emph{correct} categories but large on \emph{wrong} categories, as shown in Fig. \ref{fig:d_s_approx}. Thus we roughly let the small complexities equal to a small constant $\gamma_1$, and large complexities equal to a large $\gamma_2$. For two \emph{background} regions from different categories, suppose we have an enhanced semantic map $\mathcal{M}_e$ which is aware of their categories, then we have
\begin{equation}
D_s(R_1|\mathcal{M}_e)\sim\gamma_1|R_1|,D_s(R_2|\mathcal{M}_e)\sim\gamma_1|R_2|.
\end{equation}
where $|\cdot|$ is the cardinality. Usually, the semantic category of the union of two regions comes from the previous categories. Hence
\begin{equation}
\begin{split}
D_s(R_1\cup R_2|\mathcal{M}_e)&\sim\min{\left(\gamma_1|R_1|+\gamma_2|R_2|,\gamma_2|R_1|+\gamma_1|R_2|\right)} \\
&=\gamma_1\max(|R_1|,|R_2|)+\gamma_2\min(|R_1|,|R_2|).\\
\end{split}
\end{equation}
Let $d_s\!=\!D_s(R_i\cup R_j)\!-\!D_s(R_i)\!-\!D_s(R_j)$, then we can estimate $d_s$ with a new term
\begin{equation}
\label{eq:simd_s}
d_b\sim\left(\gamma_2-\gamma_1\right)\cdot\min\left(|R_1|,|R_2|\right).
\end{equation}

It means $d_b$ is linearly related to the minimum region size, as shown in Fig. \ref{fig:d_s_approx}.c-d. However up to now, we do not really known whether the underlying categories are different. Suppose the \emph{true} categories are different with a probability $\eta$, we use the expected value of $d_b$ instead of random guesses
\begin{equation}
\label{eq:d_s2cases}
d_b=\left\{
\begin{array}{lll}
\eta\cdot\left(\gamma_2-\gamma_1\right)\min\left(|R_1|,|R_2|\right), &{l_{R_1}=l_{R_2}=\emph{\textbf{BKG}},}\\
0, &\text{otherwise,}
\end{array} \right.
\end{equation}
where \emph{\textbf{BKG}} represents the \emph{background} category. Inspired by the recent object proposal researches \cite{Arbelaez3}, we use location-sensitive prior probabilities to capture $\eta$ in (\ref{eq:d_s2cases}), please see the supplemental materials for details. It is possible to gain more improvements if we use complicated proposal models as in \cite{Wang}.

\section{Handling the Noise in Multilevel Cues}
\label{sec:noise}

\begin{figure}
	\centering
	\begin{tabular}{ccc}
		\includegraphics[width=2.4cm]{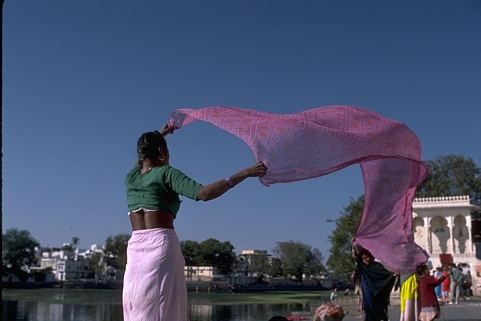}&
		\includegraphics[width=2.4cm]{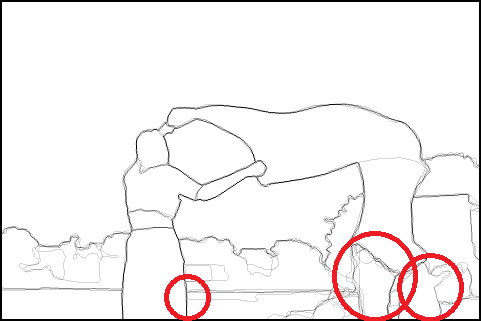}&
		\includegraphics[width=2.4cm]{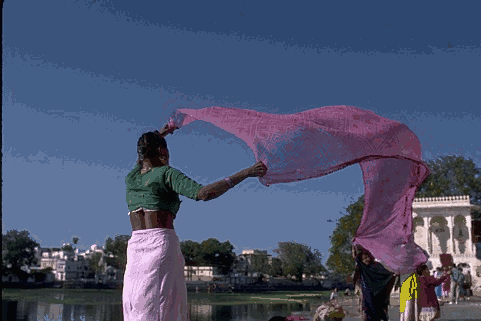}\\
		(a) & (b) & (c) \\
		\includegraphics[width=2.4cm]{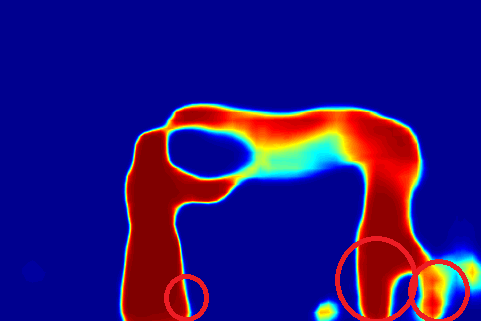}&
		\includegraphics[width=2.4cm]{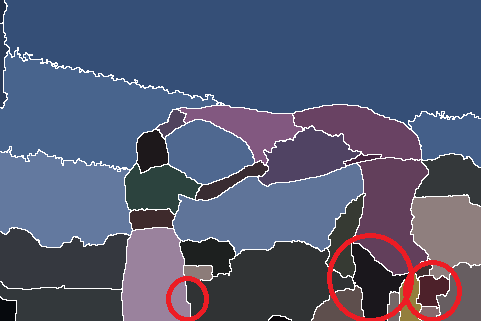}&
		\includegraphics[width=2.4cm]{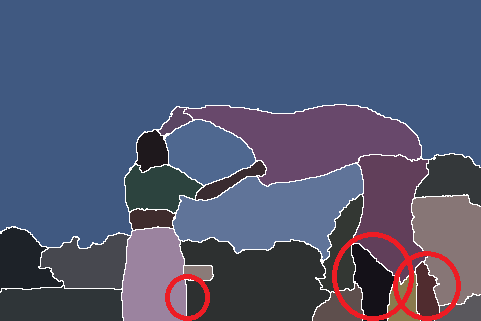}\\
		(d) & (e) & (f) \\
	\end{tabular}
	\caption{Benefit of handling noise. (a) Image. (b) Averaged contour map of multiple groundtruths. (c) Clustered colors. Pixels are shown in mean colors. (d) Semantic map of the \emph{person} category. Some noisy regions are bounded by red circles. Segmentations (e) without noise handling, and (f) with noise-handling.}
	\label{fig:noises}
\end{figure}

The noise in inaccurate semantic cues might tamper the accuracy of region boundaries as in Fig. \ref{fig:noises}.d-e. Besides this, the clustering accuracies of colors and textures have key effect to $umc$ calculations, but hard clustering methods such as K-Means might separate much similar colors/textures into different clusters. For example in Fig. \ref{fig:noises}.c, pixels in the sky region are grouped into five clusters, consequently it makes the following segmentation process divide the whole sky into several pieces. Soft clustering methods such as the Gaussian Mixture Model (GMM) use soft decision boundaries and should be capable of reducing the noise. However we find in practice its effect is far from expected. A recent work involves spatial continuity in color clustering by embedding them into manifolds, and obtains promising results \cite{Yu}. Nevertheless it is too complicated to compute easily. Finally we choose to use hard clustering, and handle the potential multilevel noise jointly. 

The cues such as inter-pixel color distances and local contours, are more accurate on locating boundaries. Intuitively, if two adjacent pixels have the same colors, they should originate from the same semantic object. So we use a logistics function with respect to the regularity term $D_r$ as a soft switch $\sigma$, to determine whether the appearance/semantic cues should weigh in:
\begin{equation}
	\label{eq:noisecoeff}
	\sigma(R_i,R_j)=\left(1+e^{-\alpha \left(\frac{d_r}{L(\{R_i,R_j\})} - \beta\right)}\right)^{-1}
\end{equation}
where $d_r=D_r(R_i\cup R_j)-D_r(R_i)-D_r(R_j)$. All these parameters will be trained. Now the unit merging cost turns into
\begin{equation}
\label{eq:handlinginsufficiency}
umc(R_i,R_j)=\frac{\sigma\left(w_cd_c+w_td_t+w_sd_s\right)+w_ud_b+w_rd_r}{L(\{R_i,R_j\})}.
\end{equation}
where $d_c,d_t$ are defined in the same way as $d_r$ and $d_s$. We do not handle the noise in $d_b$ as it is an approximation. The experiments show the noise in multilevel cues can be suppressed effectively in this way, see Fig. \ref{fig:noises}.

The trick proposed in this section makes it possible to adopt simple but fast clustering methods to extract appearance cues, meanwhile without tampering the segmentation quality much. In fact, we adopt the Principal Components Analysis (PCA) based clustering in the experiments, please see the supplementary materials for details.

\section{Experiments}
\label{sec:exp}

We aim to verify whether semantic cues can benefit image segmentations as expected, and whether we can bring some improvement to semantic segmentations in return. There are two groups of experiments: image segmentation on the BSDS500 \cite{Martin}, and semantic segmentation on the Pascal VOC12 \cite{Everingham}. All parameters are trained on the BSDS500 train set, with the strategy mixing grid searching and gradient descending as in \cite{Arbelaez1}. The semantic cues are obtained by the FCNN-based dense prediction method in \cite{Chen}. There are only 20 meaningful categories involved as the model is established for the Pascal VOC. All source code and results would be open upon acceptance for easy reproducibility.

\subsection{Image Segmentation}

\begin{table*}[htbp]
	\begin{center}
		\caption{Results on the BSDS500 test set. $\dagger$: GPU time. We use structured edges \cite{Dollar1} in \emph{Ours}, and HED \cite{Xie} (the original version with \emph{F}=0.782) in \emph{Ours-HED}. The lower VOI values are better.}
		\resizebox{12cm}{!}{ 
			\begin{scriptsize}
			\begin{tabular}{l|l|l|l|l|l|l|l|l||l|l|l||c|}
				\cline{2-13}
				\label{tab:500}
				\multirow{2}*{}
				&\multicolumn{2}{c|}{$F_{op}$}&\multicolumn{2}{c|}{Covering}&\multicolumn{2}{c|}{PRI}&\multicolumn{2}{c||}{VOI}&\multicolumn{3}{c||}{\emph{F}}&\multirow{2}*{Time}\\ \cline{2-12}
				& ODS & OIS & ODS & OIS & ODS & OIS & ODS & OIS & ODS & OIS & AP &  \\
				\hline
				\multicolumn{1}{|l|}{Human} & 0.56 & 0.56 & 0.72 & 0.72 & 0.88 & 0.88  & 1.17 & 1.17 & 0.80 & 0.80 & - & -\\
				\hline
				\multicolumn{1}{|l|}{EGB \cite{Felzenszwalb}}& 0.16 & 0.24 & 0.52 & 0.57 & 0.80 & 0.82  & 2.21 & 1.87  & 0.61 & 0.64 & 0.56 & $<$1.0s\\
				\multicolumn{1}{|l|}{NCut \cite{Shi}} & 0.21 & 0.27 & 0.45 & 0.53 & 0.78 & 0.80  & 2.23 & 1.89 & 0.64 & 0.68 & 0.45 & 600s+\\
				\multicolumn{1}{|l|}{MShift \cite{Comaniciu}}& 0.23 & 0.29 & 0.54 & 0.58 & 0.79 & 0.81  & 1.85 & 1.64 & 0.64 & 0.68 & 0.56 & 600s+\\
				\multicolumn{1}{|l|}{UCM \cite{Arbelaez1}} & 0.35 & 0.38 & 0.59 & 0.65 & 0.83 & 0.86  & 1.69 & 1.48 & 0.73 & 0.76 & 0.73 & 240s\\
				\multicolumn{1}{|l|}{ISCRA \cite{Ren}} & 0.35 & 0.42 & 0.59 & 0.66 & 0.82 & 0.85  & 1.60 & 1.42 & 0.72 & 0.75 & 0.46 & 240s+\\
				\multicolumn{1}{|l|}{MCG \cite{Arbelaez3}} & 0.38 & 0.43 & 0.61 & 0.66 & 0.83 & 0.86  & 1.57 & 1.39 & 0.75 & 0.78 & 0.76 & 20s+\\
				\hline
				\multicolumn{1}{|l|}{LEP \cite{Zhao}} & 0.42\begin{tiny}16\end{tiny} & 0.46\begin{tiny}53\end{tiny} & 0.62\begin{tiny}65\end{tiny} & 0.68\begin{tiny}55\end{tiny} & 0.83\begin{tiny}64\end{tiny} & 0.86\begin{tiny}84\end{tiny} & 1.46\begin{tiny}76\end{tiny} & 1.29\begin{tiny}13\end{tiny} & 0.75\begin{tiny}75\end{tiny} & 0.79\begin{tiny}28\end{tiny} & 0.81\begin{tiny}78\end{tiny} & 1.0s\\
				\multicolumn{1}{|l|}{Ours} & 0.43\begin{tiny}09\end{tiny} & 0.48\begin{tiny}61\end{tiny} & 0.64\begin{tiny}30\end{tiny} & 0.70\begin{tiny}12\end{tiny} & 0.84\begin{tiny}26\end{tiny} & 0.87\begin{tiny}76\end{tiny} & 1.42\begin{tiny}00\end{tiny} & 1.23\begin{tiny}19\end{tiny} & 0.75\begin{tiny}94\end{tiny} & 0.80\begin{tiny}03\end{tiny} & 0.82\begin{tiny}69\end{tiny} & 0.8s+0.4s$\dagger$\\
				\multicolumn{1}{|l|}{Ours-HED} & 0.44\begin{tiny}37\end{tiny} & 0.50\begin{tiny}14\end{tiny} & 0.66\begin{tiny}16\end{tiny} & 0.70\begin{tiny}73\end{tiny} & 0.85\begin{tiny}52\end{tiny} & 0.87\begin{tiny}85\end{tiny} & 1.35\begin{tiny}61\end{tiny} & 1.19\begin{tiny}71\end{tiny} & 0.79\begin{tiny}38\end{tiny} & 0.81\begin{tiny}60\end{tiny} & 0.84\begin{tiny}99\end{tiny} & 0.8s+0.7s$\dagger$\\
				\hline
				\hline
				\multicolumn{1}{|l|}{SE \cite{Dollar1}} & - & - & - & - & - & -  & - & - & 0.75 & 0.77 & 0.80 & 0.4s\\
				\multicolumn{1}{|l|}{HED \cite{Xie}} & - & - & - & - & - & -  & - & - & 0.78 & 0.80 & 0.83 & 0.4s$\dagger$\\
				\hline
			\end{tabular}\end{scriptsize}}
		\end{center}
	\end{table*}
	
We adopt four region-based measures including $F_{op}$ \cite{Pont-Tuset}, Covering \cite{Arbelaez1}, PRI \cite{Arbelaez1}, VOI \cite{Arbelaez1}, and a boundary-based measure $F$ \cite{Arbelaez1}. Performance is considered better on larger measures except for VOI. We collect the performance on Optimal Dataset Scales (ODS) and Optimal Image Scales (OIS) as suggested in \cite{Arbelaez1}. Our counterparts include seven methods, MCG \cite{Arbelaez3}, UCM \cite{Arbelaez1}, MShift \cite{Comaniciu}, NCut \cite{Shi}, ISCRA\cite{Ren}, EGB \cite{Felzenszwalb}, and LEP \cite{Zhao}. They are of the state-of-the-art performance or widely used nowadays.
			
\begin{figure}[htbp]
	\centering
	\begin{tabular}{cc}
		\includegraphics[width=5.05cm]{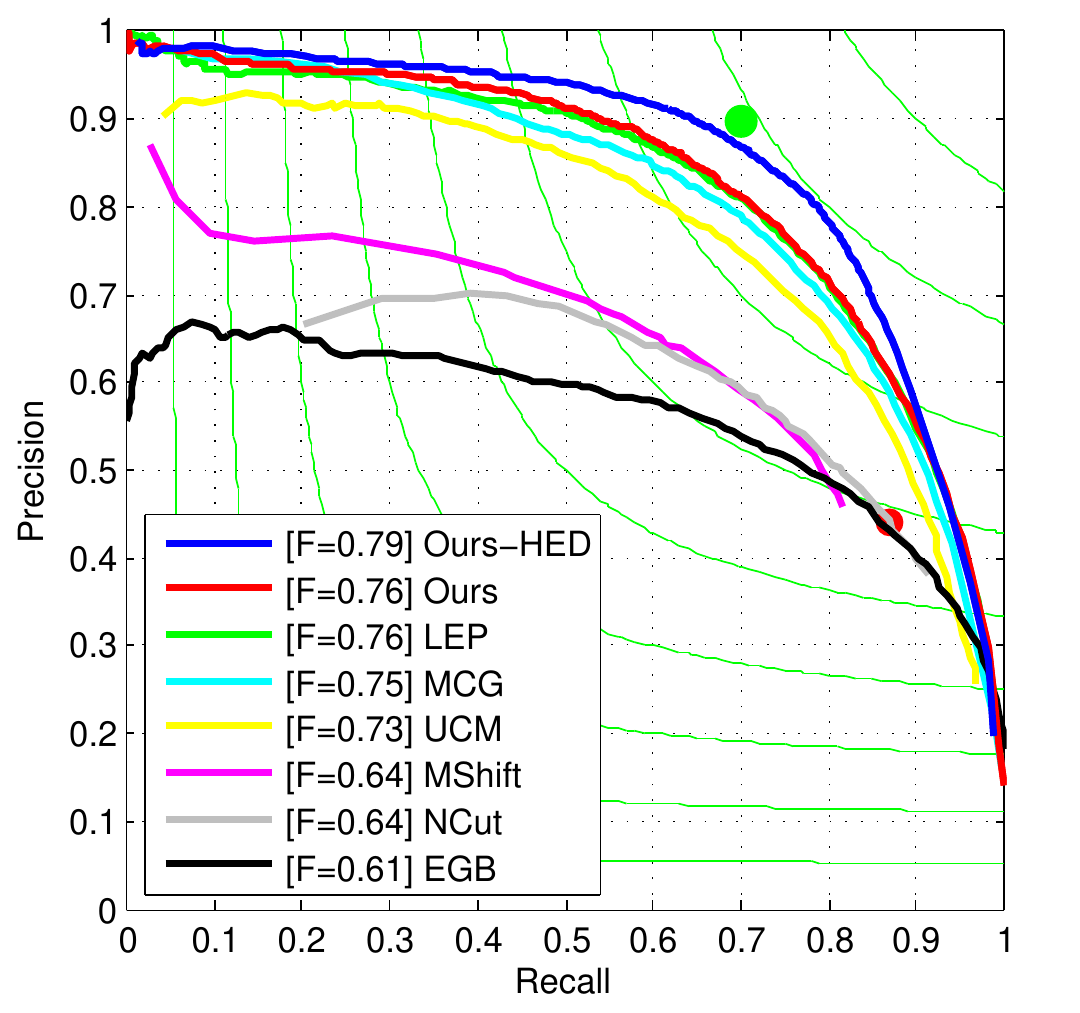}\ \ \ \ \ &
		\ \ \ \ \ \includegraphics[width=4.65cm]{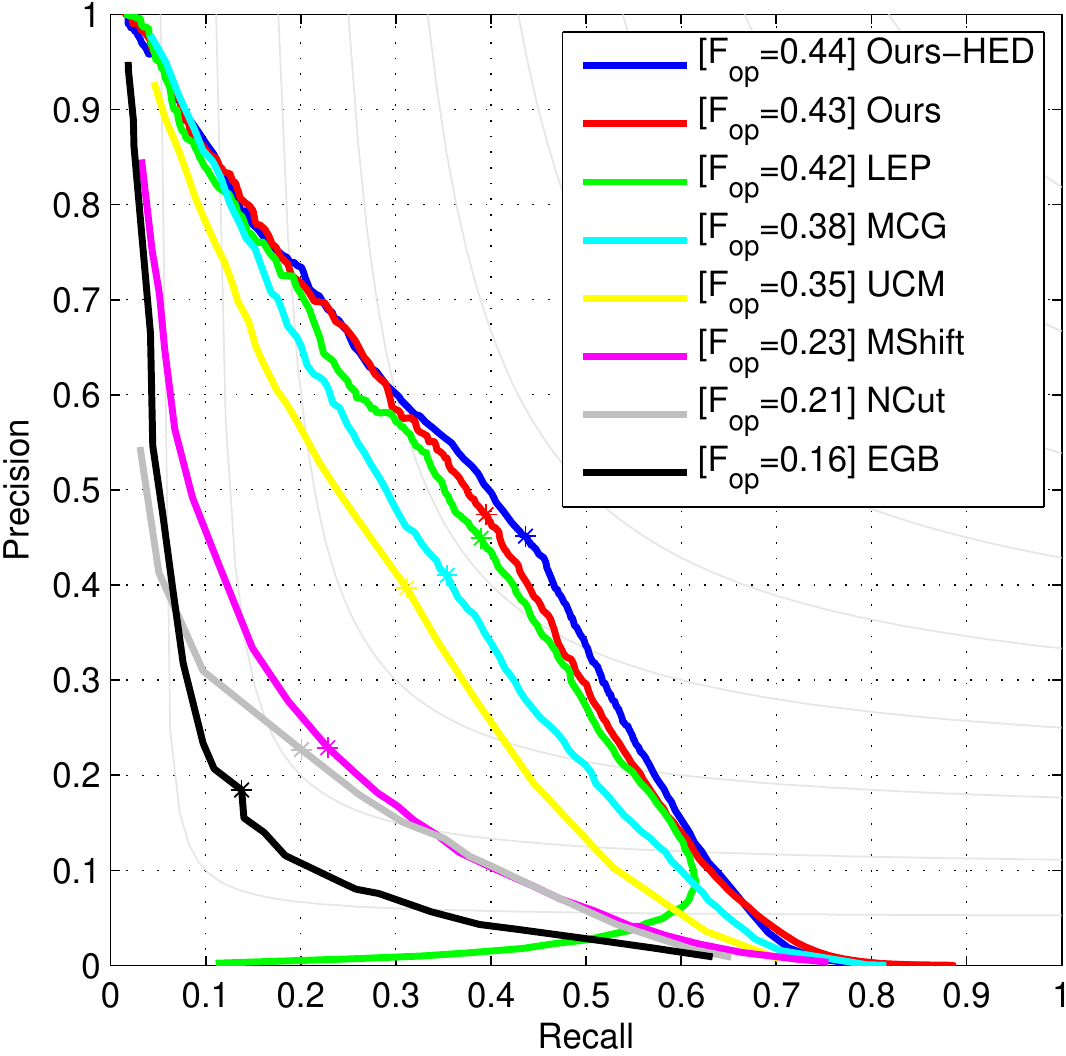}\\
		(a) & \ \ \ \ \ \ \ \ \ \ \ (b) \\
	\end{tabular}
	\caption{Performance on the BSDS500 test set. (a) \emph{F} and (b) $F_{op}$.}
	\label{fig:F}
\end{figure}

For the sake of fairness, we use the same contour cues as adopted in MCG and LEP, structured edges \cite{Dollar1}. All evaluation results are shown in Table \ref{tab:500} and Fig. \ref{fig:F}. Our method performs noticeably better on all region-based measures than its counterparts, indicating the semantic cues perform as expected to prevent merging regions of distinct categories. Especially in Fig. \ref{fig:F}.b, on the large recall rates, our $F_{op}$ values are hetergeneously better than that of LEP, indicating that the semantic cues weigh in even for small regions.

\begin{figure}[htbp]
	\centering
	\includegraphics[width=12.0cm]{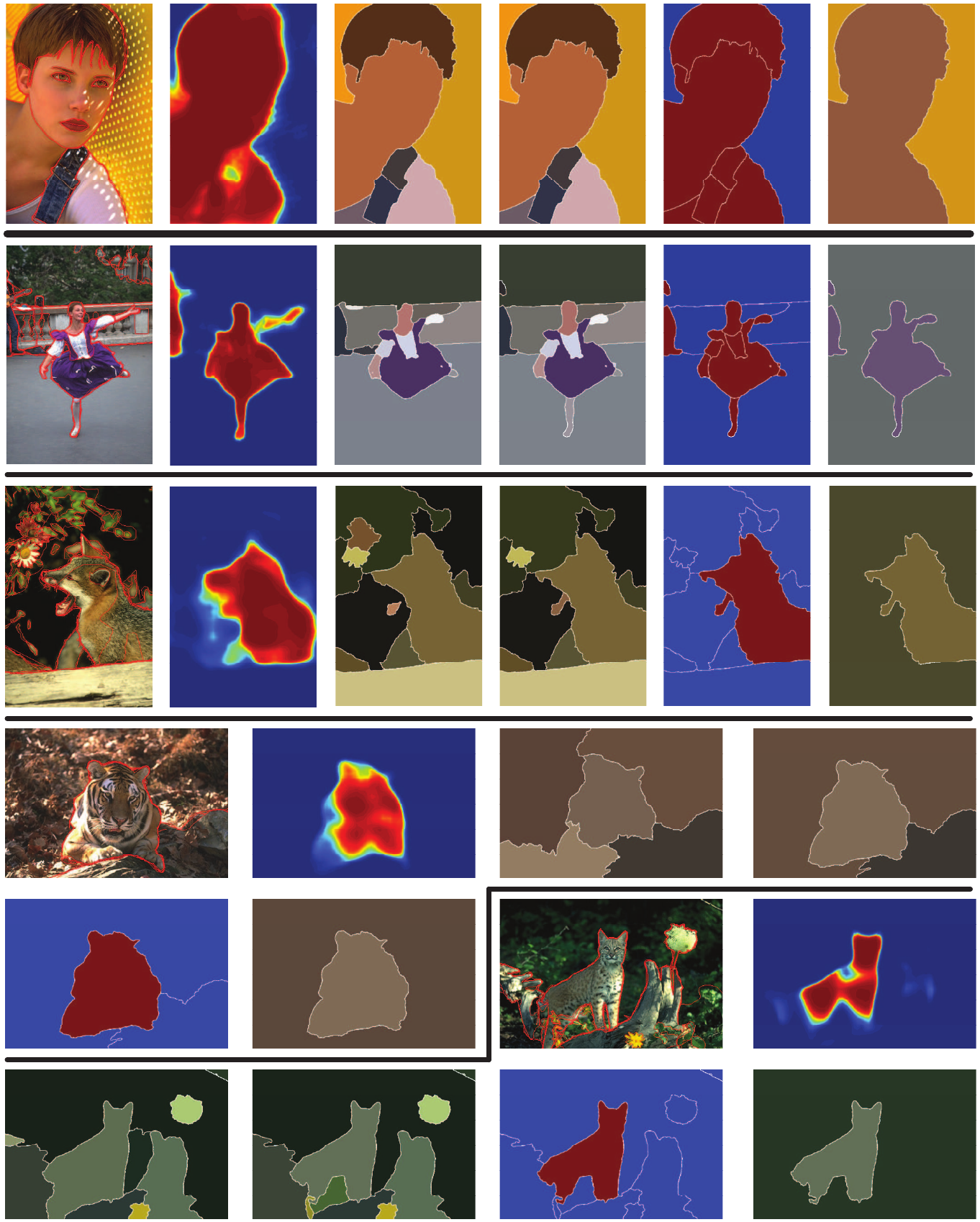}
	\caption{BSDS500 images and segmentations. In each group: image (covered by groundtruth boundaries in red), semantic map of the major category, LEP segmentation, Ours-HED segmentation, category-wise subregion map, and semantic segmentation. Note: all results are ODS ones. Zoom in for better views.}
	\label{fig:bsds500examples}
\end{figure}

On the other hand, there are only margin improvements on the boundary-based measure. It implies that semantic cues have no significant effect on contour detections, as suggested in \cite{Zitnick}. Another reason is the limited number of categories: only a small set of objects in BSDS500 have determined semantic categories. Meanwhile, some improvements are in small scales thus do not reflect much in measures. Nevertheless, they are remarkable in subjective evaluations, such as the dog mouth and the dancer leg in Fig. \ref{fig:bsds500examples}.

Interestingly, in case that semantic cues are not accurate enough, we can also obtain high quality segmentations, as in the first group and last three groups in Fig. \ref{fig:bsds500examples}. It means that semantic cues and low-/mid-level cues can correct the mistakes in their companions. Furthermore, by utilizing the cutting-edge contour cues \cite{Xie}, we find the performance on all measures can be improved remarkably. It indicates that our framework is highly flexible to accommodate cues from various sources.
			
\subsection{Semantic Segmentation}
\label{sec:semexp}
			
We use \emph{ours-HED} in this part, then compare the segmentations with the original results, and those refined with fully connected CRFs \cite{Krahenbul} in \cite{Chen}. We use the mean Intersection-over-Union (IoU) as the measure \cite{Everingham}. The threshold parameter $\lambda_\tau$ is different from that of the BSDS500 as the images have much larger sizes. We test all methods on the Pascal VOC12 val set, and the results are shown in Table \ref{tab:vocval}.
			
	\begin{table*}[htbp]
		\begin{center}
			\caption{IoUs (\%) on the Pascal VOC12 val set. The best values are shown in bold.}
			\vspace{-5.0pt}
			\resizebox{\textwidth}{!}{
				\begin{tabular}{c|c|c|c|c|c|c|c|c|c|c|c|c|c|c|c|c|c|c|c|c|c|c|}
					\cline{2-22}
					\label{tab:vocval}
					\multirow{1}*{}
					& plane & bicycle & bird & boat & bottle & bus & car & cat & chair & cow & table & dog & horse & mbike & person & plant & sheep & sofa & train & tv & mean \\
					\hline
					\multicolumn{1}{|l|}{Deeplab \cite{Chen}} & 86.7 & 58.9 & 85.8 & 76.3 & 78.4 & 91.7 & 85.7 & 87.6 & 41.3 & 88.3 & 63.4 & 84.9 & 82.0 & 81.6 & 84.3 & 65.4 & 85.6 & 63.6 & 88.0 & 68.7 & 78.2 \\
					\multicolumn{1}{|l|}{Deeplab+CRF \cite{Chen}} & 91.1 & \textbf{61.0} & 90.9 & 80.9 & 82.2 & 92.9 & 87.0 & 91.4 & 45.9 & 92.3 & \textbf{67.1} & 89.1 & 85.8 & 83.8 & 86.7 & \textbf{71.7} & 91.0 & 68.8 & \textbf{89.7} & 73.9 & 81.8 \\
					\multicolumn{1}{|l|}{Deeplab+Ours-HED} & \textbf{91.4} & 55.8 & \textbf{92.3} & \textbf{81.3} & \textbf{85.0} & \textbf{93.1} & \textbf{87.3} & \textbf{92.2} & \textbf{46.9} & \textbf{93.3} & \textbf{67.1} & \textbf{91.9} & \textbf{87.0} & \textbf{83.9} & \textbf{87.5} & 70.1 & \textbf{92.4} & \textbf{70.0} & 89.0 & \textbf{75.3} & \textbf{82.3} \\
					\hline
				\end{tabular}}
			\end{center}
		\end{table*}
		
\begin{figure}[htbp]
	\vspace{-20.0pt}
	\centering
	\includegraphics[width=12.0cm]{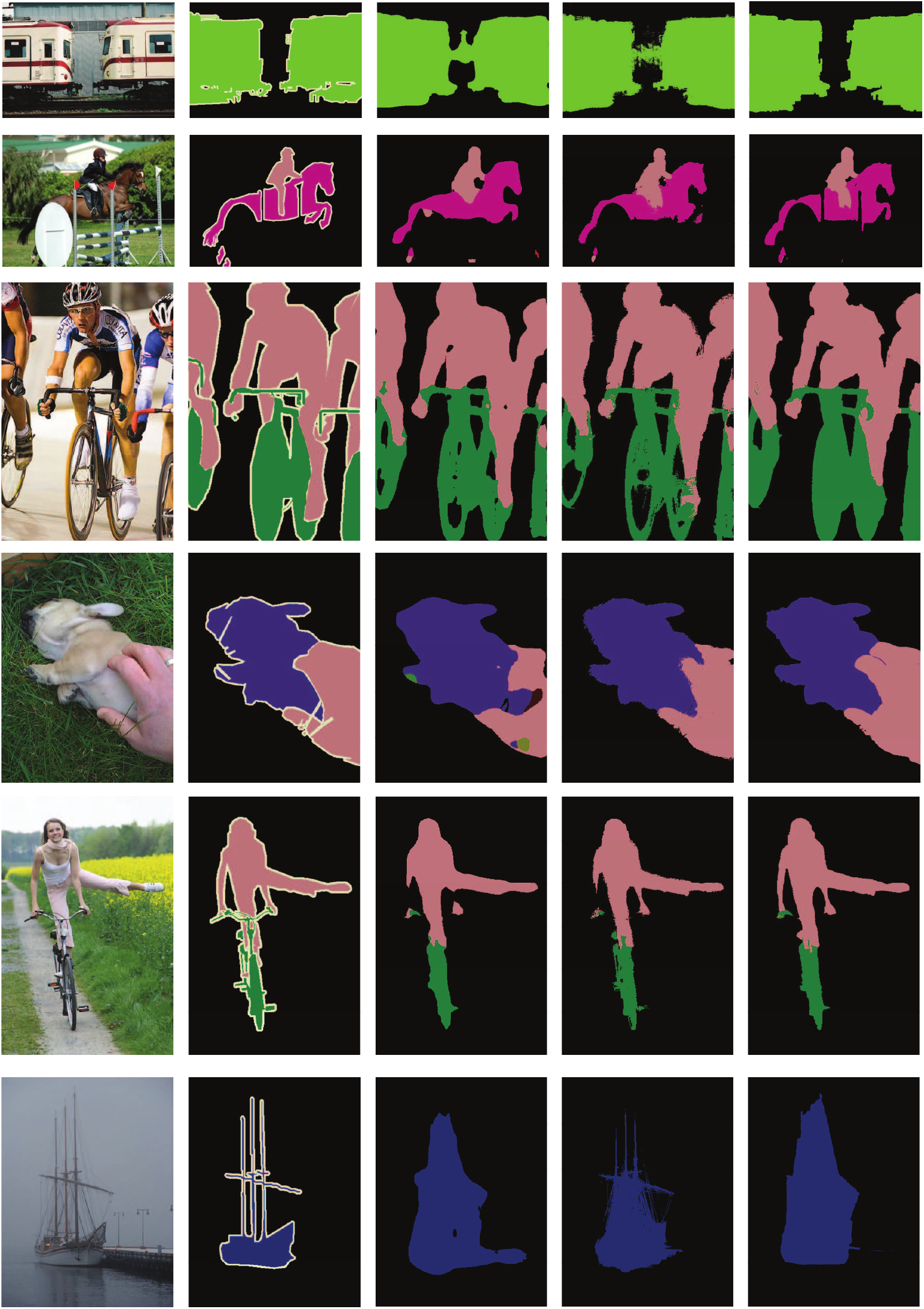}
	\caption{Pascal VOC images and segmentations. Left: image. Middle left: groundtruth. From middle to right: segmentations of Deeplab, Deeplab+CRF, and Deeplab+\emph{Ours-HED}. Note: all results are ODS ones. Zoom in for better views.}
	\label{fig:vocexamples}
\end{figure}

The performance of our method is noticeably better than that of \cite{Chen}. It proves our framework do bring some improvements to the original semantic segmentations. Especially, our advantages are more remarkable on objects of big connected regions such as bottles and dogs, however less on those of long thin structures such as the ship in Fig. \ref{fig:vocexamples}. The reason is, our method is more inclined to generate connected regions without holes, while CRF-based methods are much good at delicate structures. The style of our results looks more like that of human annotations, given that there are less noise-like structures than in CRF results. The boundaries become smoother and more accurate as expected. It makes our results more remarkable in subjective evaluations than in IoU measures, see Fig. \ref{fig:vocexamples}.
			
\section{Discussion and Conclusions}

The experiments indicate that our method obtains the better segmentation performance, and the semantic cues work as expected from the very beginning. It implies that the way we exploit semantic cues in image segmentation is effective and sensible. Besides this, the framework is flexible so that we can adopt the cues from various sources. The way we propose to handle the noise in multilevel cues is effective in improving the robustness, making it possible to adopt simple color/texture clustering methods for the higher efficiency. We can replace them with more delicate ones for the better performance. At last, we believe a more comprehensive semantic model involving plenty of categories would benefit the segmentations substantially.

Meanwhile it is easy to extract semantic segmentation results. Compared with the original results, our method gains the better performance together with more accurate region boundaries. Thus it is a good choice to refine the dense semantic segmentations.

\bibliographystyle{splncs}
\bibliography{lepbib}

\clearpage

\end{CJK*}
\end{document}